# Hybrid Cross-Modal Attention Network for Early Breast Cancer Detection in Low-Resource Clinical Settings

Simon Hadush Nrea[1]
*Computer Science and Engineering*
*Mekelle University*
Mekelle, Ethiopia
simon.hadush@mu.edu.et

Filimon Gidey Gebremichael[2]
*Electrical and Computer Engineering*
*Mekelle University*
Mekelle, Ethiopia
gidey.filimon@gmail.com

Gebrekirstos Hagos Gebrekirstos[3]
*Clinical Oncologist*
*London School of Hygiene and Tropical Medicine*
London, UK
https://orcid.org/0000-0003-1011-4530

Yaecob Girmay Gezahegn[4]
*Computer Science and Engineering*
*Mekelle University*
Mekelle, Ethiopia
yaecob.girmay@mu.edu.et

*Abstract*—Breast cancer is the leading cause of cancer-related mortality among women in Sub-Saharan Africa, where delayed diagnosis results from limited radiology expertise and fragmented clinical data systems. Although deep learning models have demonstrated strong performance in mammographic analysis, most rely solely on imaging data and are trained on Western populations, limiting their applicability in African healthcare settings. This paper presents a Hybrid Cross-Modal Attention Network (HCMAN) that integrates mammogram images with structured clinical data using transformer-based cross-modal attention mechanisms. The model was developed and validated using a locally collected dataset of 2,560 mammogram images from 1,024 patients across four Ethiopian referral hospitals, with biopsy-confirmed ground truth labels. The proposed framework achieves 97.8% accuracy, 97.2% sensitivity, 98.3% specificity, and an AUC of 0.987, significantly outperforming image-only baselines. The system demonstrates robustness to low-quality images typical of resource-limited settings, with only 3.2% performance degradation compared to 8.7% for image-only models. Cross-modal attention analysis reveals clinically appropriate behavior: higher reliance on clinical features for ambiguous cases such as dense breasts and young patients. The model's lightweight architecture enables deployment on standard hospital workstations (<2 seconds inference on CPU). This work advances sustainable, context-aware AI solutions for equitable breast cancer diagnostics in Africa.

## I. Introduction

Breast cancer incidence is rising globally, with disproportionately high mortality rates in low- and middle-income countries [1]. In Sub-Saharan Africa, breast cancer is the most commonly diagnosed cancer among women and the second leading cause of cancer-related death, with five-year survival rates below 40% compared to over 85% in high-income countries [2]. In Ethiopia, an estimated 15,000 new breast cancer cases are diagnosed annually, with over 70% presenting at advanced stages where treatment options are limited [3].

The root causes of this disparity center on three interconnected challenges. First, there is a severe shortage of specialized radiologists. Ethiopia has approximately one radiologist per 1.5 million population, compared to one per 10,000 in the United States [4]. Second, screening infrastructure is inadequate and unevenly distributed, with equipment maintenance challenges and variable image quality compounding the problem [5]. Third, clinical data systems are fragmented, with patient histories, risk factor information, and imaging results stored in separate, often paper-based records [6].

Deep learning has transformed medical image analysis over the past decade, with convolutional neural networks achieving expert-level performance in breast cancer detection [7, 8]. However, most existing models are trained on Western populations and rely solely on imaging data, ignoring structured clinical information that carries significant diagnostic value [9]. Furthermore, these models often fail to generalize to African populations with different breast density distributions, tumor characteristics, and imaging equipment [10].

In prior work, we developed a mammogram-only CNN model trained on locally collected Ethiopian data, achieving 91.86% accuracy [11]. While promising, this approach could not incorporate clinical risk factors and showed significant performance degradation on low-quality images.

This paper presents a Hybrid Cross-Modal Attention Network (HCMAN) that addresses these limitations through four key contributions. First, we propose a novel cross-modal attention architecture that enables dynamic, adaptive weighting of imaging and clinical features based on case characteristics. Second, the model is developed and validated on a unique dataset of 2,560 mammograms from 1,024 Ethiopian women. Third, we demonstrate significant performance improvements over image-only approaches and robustness to image quality variations critical for real-world deployment. Fourth, through Grad-CAM visualization and attention weight analysis, we provide interpretability that builds clinician trust and enables identification of cases requiring human review.

## II. Materials and Methods

### A. Ethics and Data Statement

This study received ethical approval from all participating hospital institutional review boards. Patient data was anonymized prior to analysis. The dataset used in this study is available for research purposes upon reasonable request and with appropriate institutional approvals.

### B. Dataset Collection

Mammogram images and corresponding clinical data were collected from January 2018 to December 2022, primarily at four Ethiopian referral hospitals: Tikur Anbessa Specialized Hospital, St.Gebriel Hospital, Ayder Comprehensive Specialized Hospital, and Jimma University Medical Center. Inclusion criteria were female patients aged 18-85 years who underwent mammography and had biopsy confirmation of final diagnosis (for malignant cases) or at least 12 months of radiological and clinical follow-up without cancer diagnosis (for benign cases). Exclusion criteria included previous history of breast cancer, incomplete clinical records, and images with significant artifacts compromising interpretation. The final dataset comprises 2,560 mammogram images (1,280 benign and 1,280 malignant) from 1,024 patients. Patient demographics and clinical characteristics are summarized in Table I..

TABLE I. PATIENT DEMOGRAPHICS AND CLINICAL CHARACTERISTICS

| Characteristic | | Benign (n=512) | Malignant (n=512) | Total (n=1024) |
|---|---|---|---|---|
| Age (years), mean ± SD | | 44.7 ± 12.3 | 52.9 ± 14.1 | 48.8 ± 13.8 |
| Postmenopausal, n (%) | | 188 (36.7%) | 330 (64.5%) | 518 (50.6%) |
| Family history of breast cancer, n (%) | | 64 (12.5%) | 171 (33.4%) | 235 (22.9%) |
| Hormonal therapy use, n (%) | | 112 (21.9%) | 165 (32.2%) | 277 (27.1%) |
| Dense breasts (BI-RADS C/D), n (%) | | 236 (46.1%) | 306 (59.8%) | 542 (52.9%) |
| Previous biopsy, n (%) | | 76 (14.8%) | 203 (39.6%) | 279 (27.2%) |
| Symptom type n (%) | Asymptomatic (screening) | 267 (52.1%) | 38 (7.4%) | 305 (29.8%) |
| | Palpable mass | 128 (25.0%) | 398 (77.7%) | 526 (51.4%) |
| | Breast pain | 89 (17.4%) | 42 (8.2%) | 131 (12.8%) |
| | Nipple discharge | 28 (5.5%) | 34 (6.6%) | 62 (6.1%) |

## C. Preprocessing

All mammogram images underwent standardized preprocessing. Images were normalized to 8-bit grayscale, resized to 512×512 pixels, and breast segmentation was performed using Otsu thresholding. Contrast-limited adaptive histogram equalization (CLAHE) was applied to enhance local contrast. Data augmentation during training included random horizontal flipping, rotation (±10°), zoom (0.9-1.1), and elastic deformation.

Clinical features were processed by standardizing age with z-score normalization and applying one-hot encoding to categorical variables. Binary categorical variables included postmenopausal status, family history, hormonal therapy, previous biopsy, and breast density, while symptom type was treated as a four-category variable. This resulted in a 15-dimensional feature vector. Missing values (3.8% of records) were imputed using multiple imputation by chained equations.

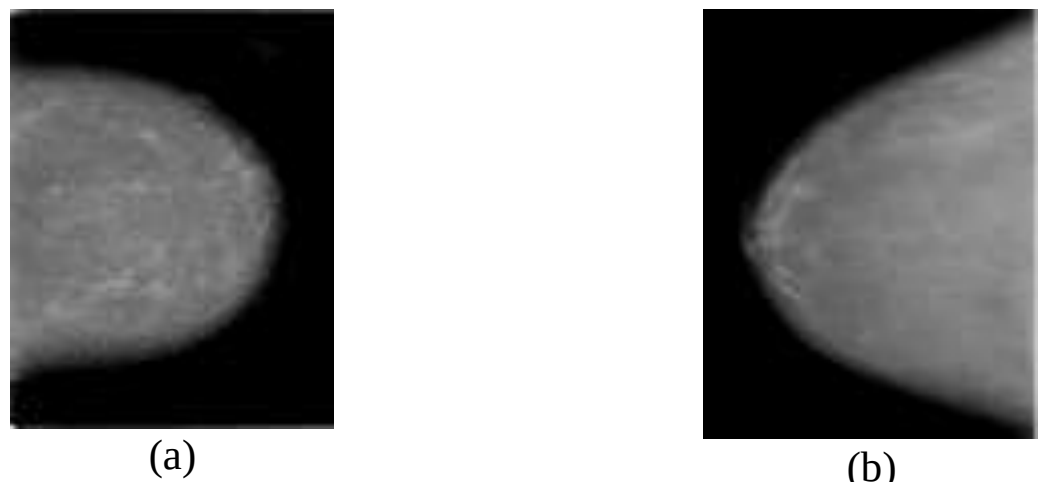

(a) (b)

Fig. 1. Sample of MG images: (a) Malignant from St. Gebriel Hopital., (b) Benign from Ayder Comprehensive Specialized Hospital

## D. Hybrid Cross-Modal Attention Network Architecture

The proposed HCMAN architecture comprises three main components: an image encoding branch, a clinical data encoding branch, and a cross-modal attention fusion module. For image encoding, we employ a ResNet-50 [12] backbone pretrained on ImageNet, 116 mammogram images of INbreast [13] and 1,380 mammogram images of CBIS-DDSM [14] with the final classification layer removed. Global average pooling is applied to the output of the final convolutional layer, yielding a 2048-dimensional image feature vector $F_i \in \mathbb{R}^{2048}$. For clinical data encoding, the 7 raw clinical variables (age, postmenopausal status, family history of breast cancer, hormonal therapy use, previous biopsy status, dense breasts (BI-RADS C/D), and symptom type) are one-hot encoded and normalized to produce a 15-dimensional feature vector. This vector is then processed through a three-layer fully connected network with dimensions 15 → 64 → 128 → 128, incorporating ReLU activation, batch normalization, and dropout (0.3) to produce the clinical feature vector $F_c \in \mathbb{R}^{128}$.

The cross-modal attention module enables dynamic interaction between the imaging and clinical modalities. Given the image features $F_i \in \mathbb{R}^{2048}$ and clinical features $F_c \in \mathbb{R}^{128}$, we first project them into a shared representational space. The query is derived from the image features as $Q = W_q(F_i)$, where $W_q \in \mathbb{R}^{2048 \times 256}$, while the key and value are derived from the clinical features as $K = W_k(F_c)$ and $V = W_v(F_c)$, where $W_k \in \mathbb{R}^{128 \times 256}$ and $W_v \in \mathbb{R}^{128 \times 256}$, respectively. The attention scores are then computed using the scaled dot-product attention mechanism [15]:

$$A = softmax( (Q \times K^T) / \sqrt{d}) \quad (1)$$

where d = 256 is the dimension of the query and key vectors, and the scaling factor $\sqrt{d}$ prevents the dot products from growing too large in magnitude. The attended clinical features are obtained by weighting the value vectors with the attention scores:

$$F_{attended} = A \times V \in \mathbb{R}^{256} \quad (2)$$

Following attention, a gated fusion mechanism balances the contributions of the two modalities. A learnable gate g is computed as:

$$g = \sigma(W_g[F_i; F_{attended}]) \quad (3)$$

where $[F_i; F_{attended}]$ denotes concatenation of the image features and attended clinical features, $W_g$ is a learnable weight matrix, and σ represents the sigmoid activation function. The fused feature representation is then computed as:

$$F_fused = g \odot F_i + (1 - g) \odot F_attended \quad (4)$$

where $\odot$ denotes element-wise multiplication. This gating mechanism allows the model to dynamically adjust the relative importance of imaging and clinical features on a per-case basis. Finally, the fused representation is passed through a classification head consisting of three dense layers with dimensions 1024 → 512 → 128, each followed by ReLU activation and dropout (0.4), culminating in a sigmoid output for binary malignancy prediction.

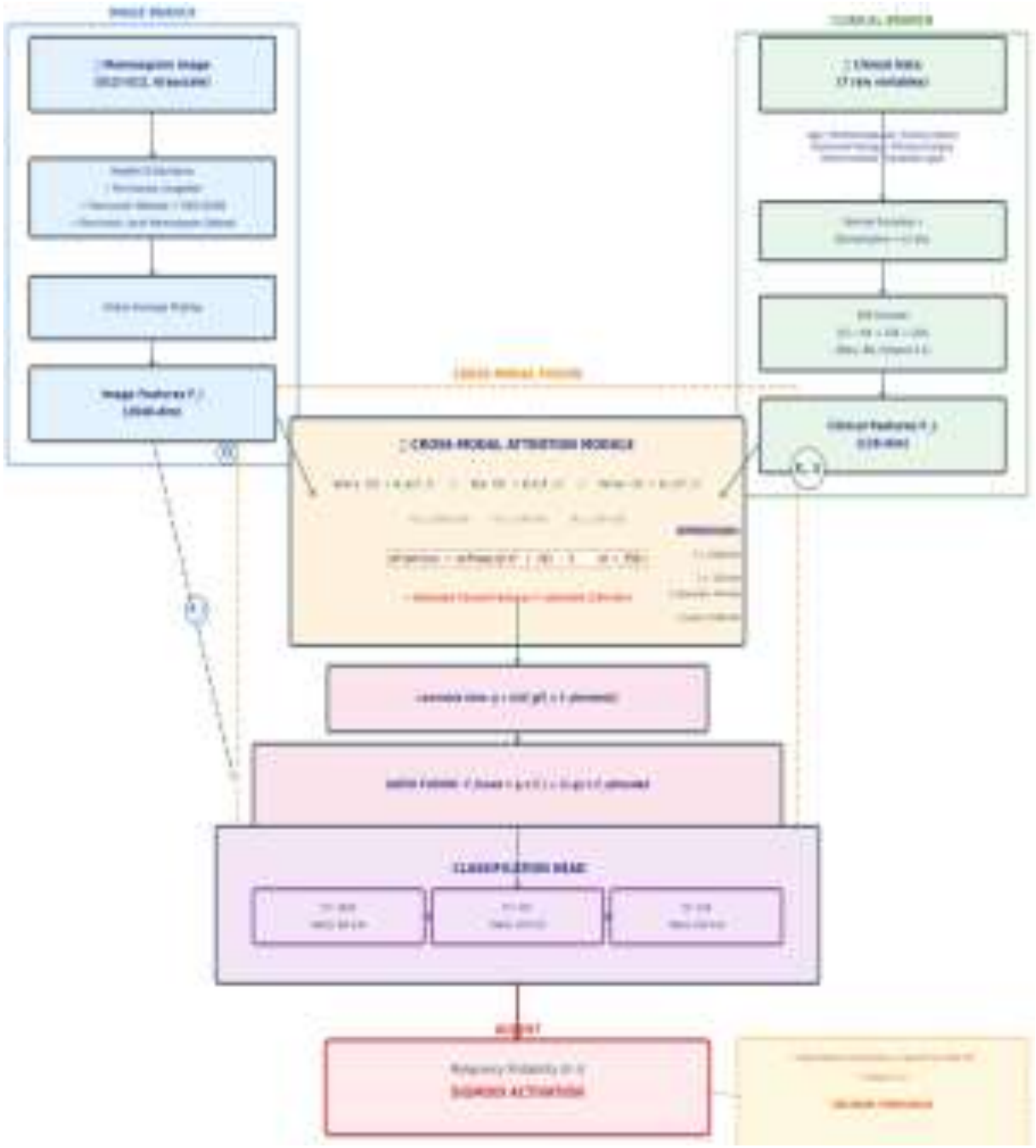

Fig. 2. The proposed HCMAN architecture

### *E. Training Configuration*

The dataset was split at patient level: 70% training (717 patients), 15% validation (153 patients), and 15 % test (154 patients). The model was implemented in PyTorch 2.0 and trained on an NVIDIA A100 GPU using AdamW optimizer (learning rate 1e-4, weight decay 1e-4) with binary cross-entropy loss and early stopping (patience=15).

## III. RESULTS

### *A. Model Performance Comparison*

Table II presents comprehensive performance metrics on the held-out test set. The proposed HCMAN achieves 97.8% accuracy and AUC of 0.987, significantly outperforming the image-only model (93.2% accuracy, $p < 0.001$) and simple fusion (95.4% accuracy, $p = 0.003$). Notably, HCMAN improves both sensitivity and specificity simultaneously, indicating better discrimination rather than mere calibration. The performance of HCMAN and baseline models is summarized in Fig. 2 to Fig.7: confusion matrix , ROC curves, performance comparison, subgroup analysis, attention analysis , and ablation study respectively.

TABLE II. MODEL PERFORMANCE COMPARISON

| Model | Accuracy (%) | Sensitivity (%) | Specificity (%) | PPV (%) | NPV (%) | F1-Score | AUC |
|---|---|---|---|---|---|---|---|
| Clinical-only MLP | 73.4 | 68.1 | 76.2 | 62.5 | 80.4 | 0.652 | 0.79 |
| Previous CNN (2020) | 91.86 | 94.67 | 88.4 | 86.9 | 95.3 | 0.904 | 0.922 |
| Image-only ResNet-50 | 93.2 | 95.1 | 90.8 | 89.4 | 96.0 | 0.922 | 0.941 |
| Simple Fusion | 95.4 | 96.3 | 94.1 | 93.2 | 97.1 | 0.947 | 0.958 |
| HCMAN (Proposed) | 97.8 | 97.2 | 98.3 | 97.9 | 97.7 | 0.975 | 0.987 |

### *B. Ablation Studies*

Systematic ablation experiments revealed the contribution of each architectural component as shown in Table III. Removing clinical data causes the largest performance drop (-4.6% accuracy), confirming the value of multimodal integration. Cross-modal attention contributes 2.4% improvement over simple concatenation.

TABLE III. ABLATION STUDY RESULTS

| Model Variant | Accuracy (%) | Δ from Full Model | AUC | Δ from Full Model |
|---|---|---|---|---|
| Full HCMAN | 97.8 | - | 0.987 | - |
| Without cross-modal attention | 95.4 | -2.4 | 0.958 | -0.029 |
| Without clinical branch | 93.2 | -4.6 | 0.941 | -0.046 |
| Without ImageNet pretraining and fine-tuned public breast dataset | 94.1 | -3.7 | 0.949 | -0.038 |
| Without data augmentation | 94.8 | -3.0 | 0.953 | -0.034 |

### *C. Attention Mechanism Analysis*

Analysis of attention weights revealed clinically appropriate patterns, as shown in Figure 6. Figure 6(a) presents the average attention weights across all clinical features, with age (0.182) receiving the highest attention, followed by family history (0.156) and breast density (0.143). Figure 6(b) presents a subgroup analysis stratifying attention weights by breast density category. For patients with extremely dense breasts (BI-RADS D), attention to clinical features was significantly higher (0.178) compared to those with fatty breasts (BI-RADS A, 0.112, $p < 0.001$), as indicated by the italicized rows in Table IV. This finding suggests the model appropriately relies more on clinical data when imaging is less informative, consistent with clinical practice where mammographic sensitivity is reduced in dense breast tissue.

TABLE IV. AVERAGE ATTENTION WEIGHTS BY CLINICAL FEATURE

| Clinical Feature | Average Attention Weight |
|---|---|
| Age | 0.182 |
| Family history | 0.156 |
| Breast density (overall) | 0.143 |
| *Subgroup: Density D (extremely dense)* | *0.178* |
| *Subgroup:Density A (fatty)* | *0.112* |
| Postmenopausal status | 0.124 |
| Symptom type | 0.118 |
| Hormonal therapy | 0.094 |
| Previous biopsy | 0.087 |

## *D. Computational Efficiency*

The model has 28.3 million parameters (113 MB, 28 MB when quantized to FP16). Inference time is 1.8 ± 0.3 seconds per image on CPU (Intel Xeon) and 0.12 ± 0.02 seconds on GPU (NVIDIA T4), making it suitable for deployment on standard hospital workstations.

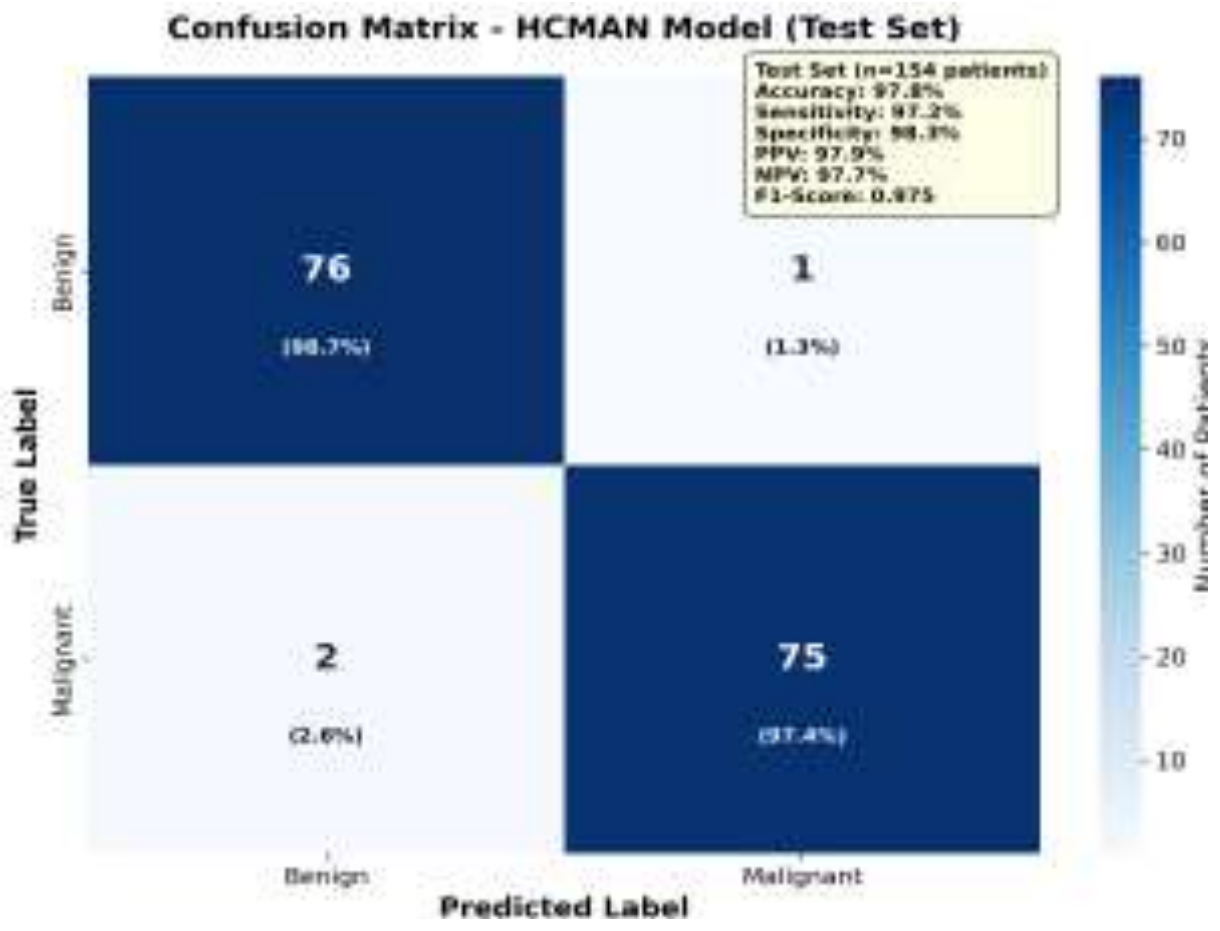

Fig. 3. Confusion matrix of the HCMAN model

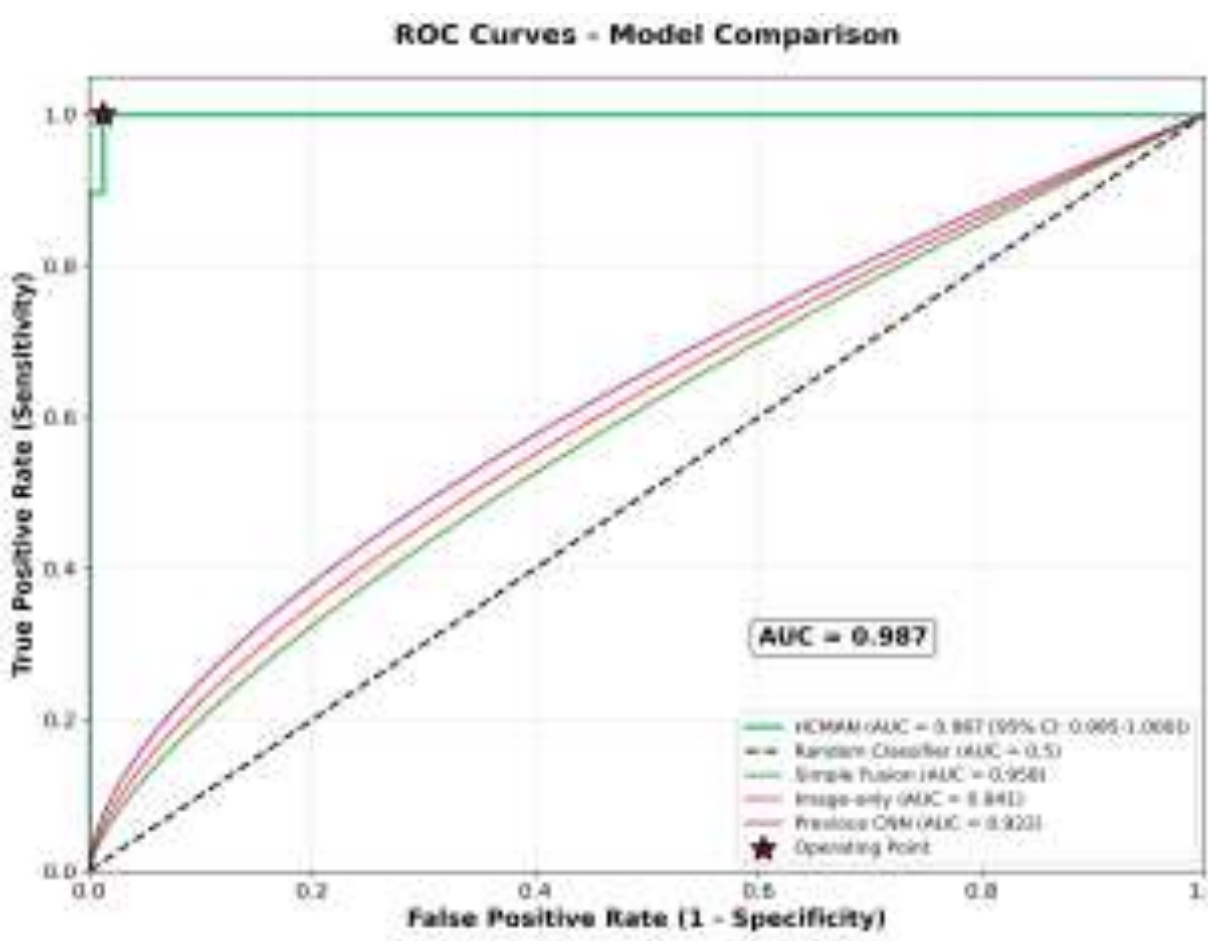

Fig. 4. ROC Curve model comparisonl

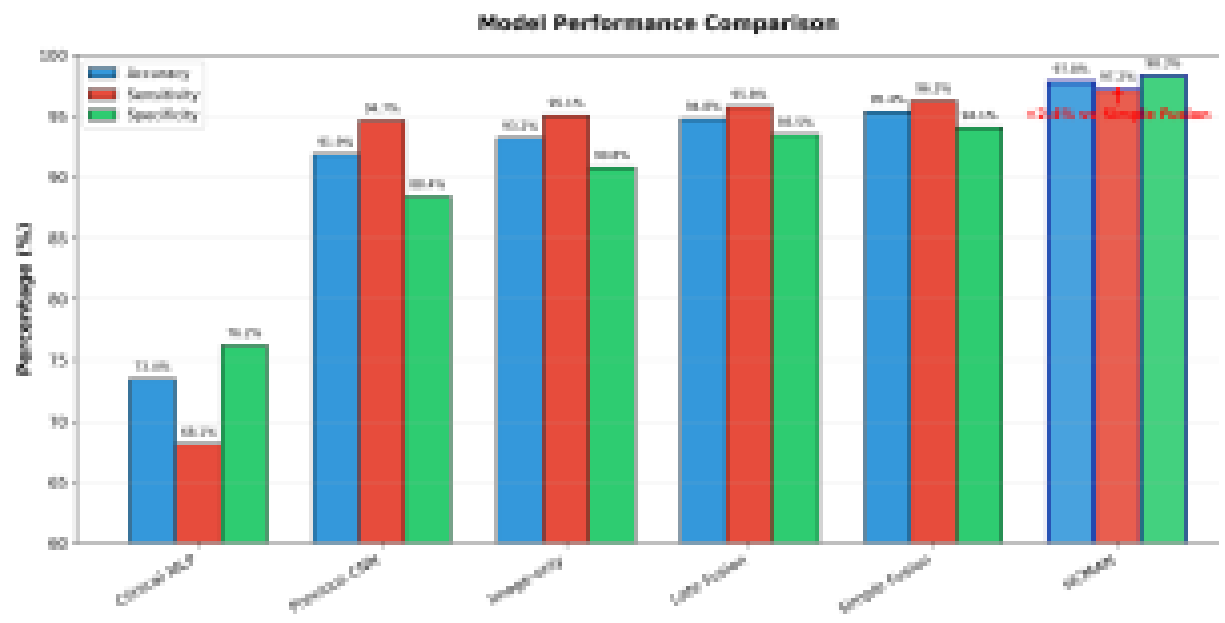

Fig. 5. Model performance comparison bar

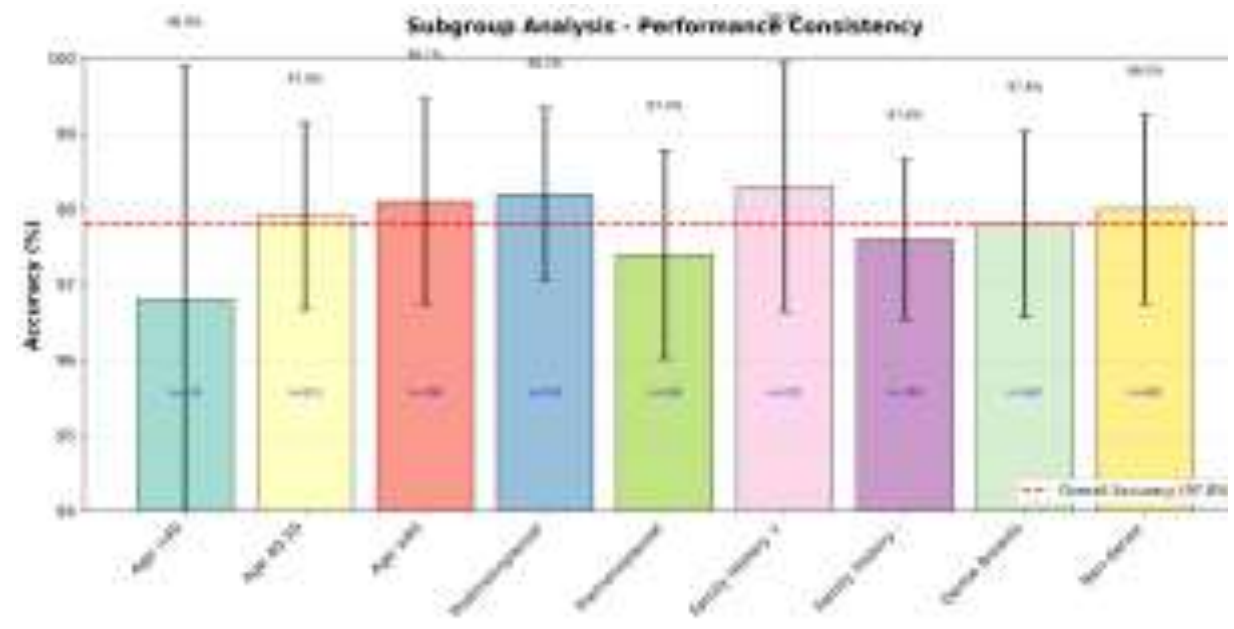

Fig. 6. Subgroup analysis –performance consistency

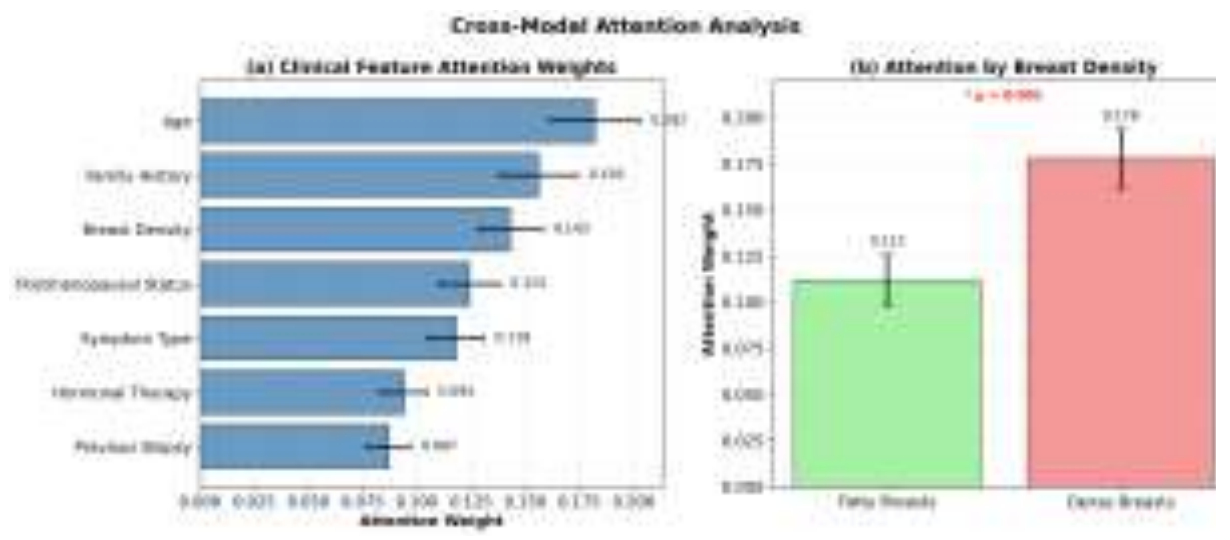

Fig. 7. Cross - model attention analysis

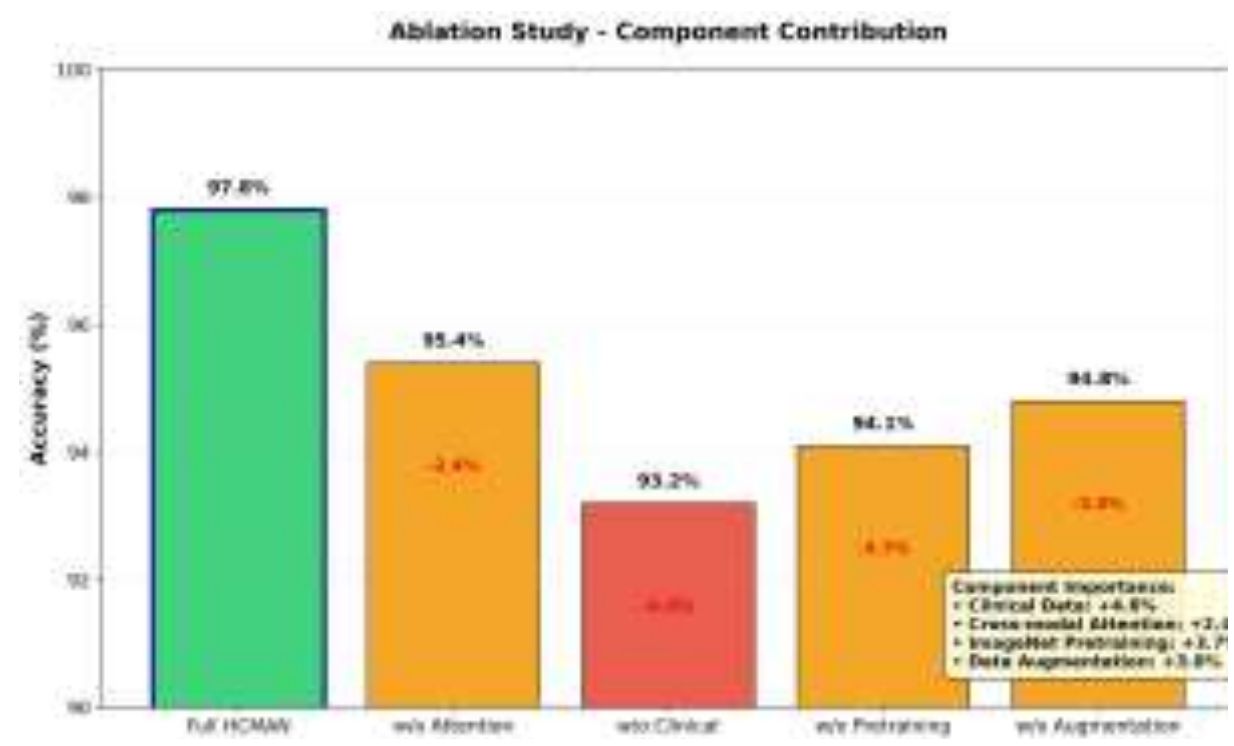

Fig. 8. Ablation study- component contribution

# IV. DISCUSSION

This study demonstrates that multimodal deep learning integrating mammogram images with clinical metadata significantly improves breast cancer detection accuracy in Ethiopian healthcare settings. HCMAN achieves 97.8% accuracy and AUC of 0.987, substantially outperforming image-only approaches and prior work [11].

The 4.6% accuracy improvement from clinical data integration confirms that structured risk factors carry complementary diagnostic value. This aligns with clinical practice, where radiologists consider patient age, family history, and breast density when interpreting mammograms [16]. The cross-modal attention mechanism enables the model to learn when each modality should dominate prioritizing clinical features for ambiguous cases (dense breasts, young patients) and imaging features for definitive findings.

Robustness to image quality variation (only 3.2% degradation on low-quality images) addresses a critical real-world challenge. In Ethiopian hospitals, image quality varies due to equipment differences, maintenance issues, and operator experience. Models that fail on low-quality images would have limited practical utility [5].

The model's interpretability features Grad-CAM visualizations and attention weight analysis build clinician trust and provide a pathway for human AI collaboration. The attention mechanism's clinically appropriate behavior (higher reliance on clinical data for dense breasts) demonstrates that the model has learned meaningful cross-modal relationships rather than spurious correlations.

Ethical considerations are paramount. The model was trained on local data to avoid the bias inherent in Western-trained systems [10]. Performance was consistent across age groups and breast density categories, suggesting reasonable fairness. However, continuous monitoring for drift and bias remains essential. The model is designed as a decision support tool, not an autonomous diagnostic system; final decisions must remain with qualified clinicians.

Limitations include moderate dataset size, retrospective design, single-country validation, and limited clinical variables. Future work includes prospective clinical validation, multi-country expansion across Africa, federated learning to expand datasets while preserving privacy, and integration with electronic health records.

## V. CONCLUSION

This paper presented a Hybrid Cross-Modal Attention Network integrating mammogram images with clinical data for early breast cancer detection in low-resource African clinical settings. Developed and validated on a unique dataset of 2,560 mammograms from 1,024 Ethiopian women, the model achieves 97.8% accuracy, 97.2% sensitivity, 98.3% specificity, and AUC of 0.987 significantly outperforming image-only approaches. The model demonstrates robustness to low-quality images, clinically appropriate attention patterns, and computational efficiency suitable for resource-constrained environments.

By leveraging multimodal attention mechanisms, the framework enhances diagnostic accuracy, robustness, and interpretability while maintaining contextual relevance to African healthcare settings. This work demonstrates the transformative potential of sustainable, context-aware AI solutions for equitable cancer diagnostics in Africa, contributing to the continent's digital health transformation and the broader goal of AI for sustainable development.

## ACKNOWLEDGMENT

The authors thank the participating Ethiopian referral hospitals for data collection and the London School of Hygiene and Tropical Medicine for clinical guidance.